\documentclass{article}

\usepackage{PRIMEarxiv}

\usepackage[utf8]{inputenc} 
\usepackage[T1]{fontenc}    
\usepackage{hyperref}       
\usepackage{url}            
\usepackage{booktabs}       
\usepackage{amsfonts}       
\usepackage{nicefrac}       
\usepackage{microtype}      
\usepackage{lipsum}
\usepackage{fancyhdr}       
\usepackage{graphicx}       
\graphicspath{{media/}}     
\usepackage{algorithm}
\usepackage{algorithmic}
\usepackage{stackengine}
\newcommand\xrowht[2][0]{\addstackgap[.5\dimexpr#2\relax]{\vphantom{#1}}}
\hypersetup{%
    pdfborder = {0 0 0}
}

\title{Reinforcement Learning-Based Automatic Berthing System
}

\author{
  Daesoo Lee \\
  Norwegian University of Science and Technology
}

\begin{document}
\maketitle

\begin{abstract}
Previous studies on automatic berthing systems based on artificial neural network (ANN) showed great berthing performance by training the ANN with ship berthing data as training data. However, because the ANN requires a large amount of training data to yield robust performance, the ANN-based automatic berthing system is somewhat limited due to the difficulty in obtaining the berthing data. In this study, to overcome this difficulty, the automatic berthing system based on one of the reinforcement learning (RL) algorithms, proximal policy optimization (PPO), is proposed because the RL algorithms can learn an optimal control policy through trial-and-error by interacting with a given environment and does not require any pre-obtained training data, where the control policy in the proposed PPO-based automatic berthing system controls revolutions per second (RPS) and rudder angle of a ship. Finally, it is shown that the proposed PPO-based automatic berthing system eliminates the need for obtaining the training dataset and shows great potential for the actual berthing application.
\end{abstract}


\section{Introduction}
Ship berthing was traditionally conducted by an experienced captain because when a ship approaches a port or harbor, the ship experiences nonlinear ship motions due to its slow speed and sudden changes in a rudder angle and engine revolutions per second (RPS) and these factors made automation of the berthing process difficult. To overcome this limitation, many studies on automatic berthing systems have been conducted and an ANN-based automatic berthing system has shown the most robust performance \cite{bae2008study, ahmed2013automatic, im2018artificial, lee2020application}. Although the previous studies on the ANN-based automatic berthing systems showed robust berthing performance, the berthing performance was still limited by a number of training data. Typically, for a well-trained ANN system, a large number of training data is required. However, in case of the ANN-based berthing system, it is hard to obtain a large number of berthing data as training data, therefore, the performance of the trained ANN in the ANN-based automatic berthing system is limited. In this study, to overcome this limitation, a proximal policy optimization (PPO)-based automatic berthing system is proposed in which the PPO is one of the most popular reinforcement learning (RL) algorithms \cite{schulman2017proximal}. The RL algorithms can train a neural network to take appropriate actions to achieve a given goal by finding an optimal control policy through interaction with a given environment. In the proposed PPO-based automatic berthing system, types of the actions are controls over the rudder angle and RPS, and the goal is to arrive at a berthing goal point. Through the interaction with a given environment, the proposed PPO-based automatic berthing system may learn to provide optimal controls over the rudder angle and RPS, which results in robust berthing from various initial positions. Finally, it is shown that the proposed PPO-based automatic berthing system can train a neural network to arrive at the berthing goal point without any pre-obtained berthing data as training data. In the following sections, a reinforcement learning theory, details of the proposed PPO-based automatic berthing systems are presented. Then, the simulation conditions and simulation and result discussion are followed.

\section{Reinforcement Learning Theory}
\begin{figure}[t]
\centering
\includegraphics[width=0.5\columnwidth]{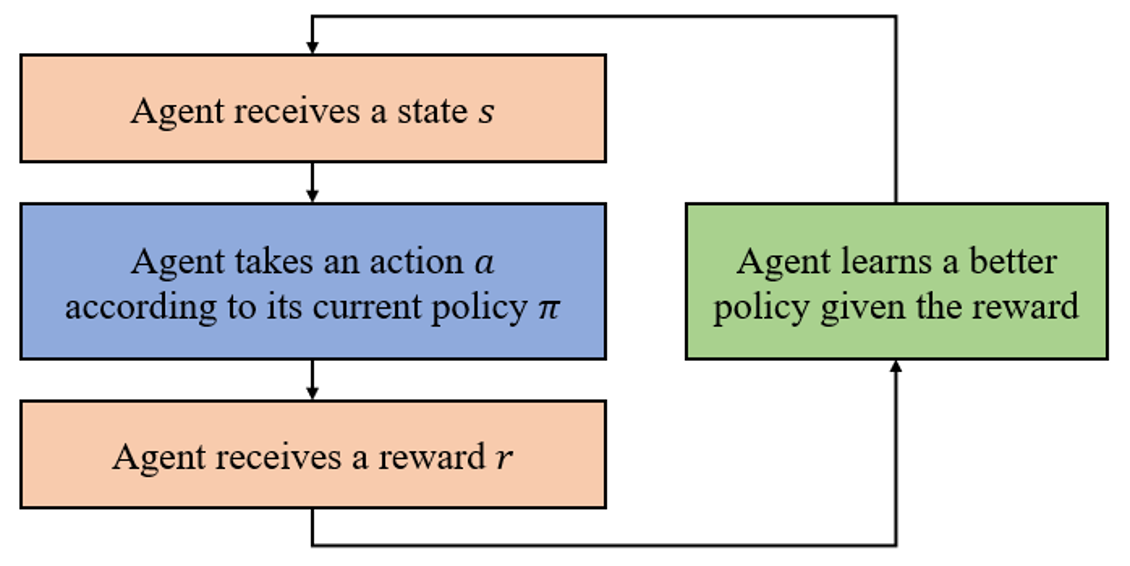}
\caption{Procedure of RL training according to MDP.}
\label{fig:procedure_of_RL_MDP}
\end{figure}

The RL is applied to problems where decisions for actions are made in a sequential order such as $\left(\mathrm{state}_t, \mathrm{action}_t, \mathrm{reward}_{t+1}, \mathrm{state}_{t+1},\cdots\right)$. A unit that learns through interaction with an environment $E$ is called an agent. To solve the sequential-action-decision problems, the problem should be defined mathematically, which can be achieved by a Markov Decision Process (MDP). The MDP consists of the state $s$, action $a$, reward $r$, action (control) policy $\pi$ in which the policy $\pi$ is a policy for taking the action $a$ given the state $s$. The procedure of the RL training according to the MDP is shown in Fig. \ref{fig:procedure_of_RL_MDP}. As the repeating loop in Fig. \ref{fig:procedure_of_RL_MDP} continues, the policy continues being updated to maximize the reward and eventually the optimal policy is achieved that can maximize the reward. Therefore, solving the MDP is what the RL algorithm does to find the optimal policy.

In the early days of the RL, the MDP was solved by a dynamic programming where the states, rewards, policy are calculated using table forms and a perfectly defined environment is required. However, in the real world, environments are very complex and involve non-linearity. Thus, it was impossible to learn a mapping function for the optimal policy using the dynamic programming for real-world applications. But, in recent years, the RL algorithms became capable of learning the optimal policy for real-world applications with the help of the ANN or other types of neural networks in which such neural networks act as an approximate mapping function for the policy $\pi$. The PPO is one of the RL algorithms to train such neural networks to map the policy efficiently and it has shown robust performance in various applications. The overall procedure of RL training with the PPO is shown in Algorithm \ref{alg:overall_procedure_PPO_theory} where the agent is basically composed of the neural networks.

\begin{algorithm}[h]
\caption{Overall procedure of RL training with PPO.}
\label{alg:overall_procedure_PPO_theory}
\begin{algorithmic}[1] 
\STATE Set an initial state $s_0$.
\FOR{$t = 0.1, 0.2, \cdots$}
    \STATE Agent takes action $a_t$ given the current state $s_t$.
    \STATE Interact with an environment.
    \STATE Obtain reward $r_{t+1}$ and next state $s_{t+1}$.
    \STATE 
    \STATE Every $N$ steps, train the agent using PPO.
\ENDFOR
\end{algorithmic}
\end{algorithm}

\section{Proposed PPO-Based Automatic Berthing System}
\begin{figure}[h]
\centering
\includegraphics[width=0.3\columnwidth]{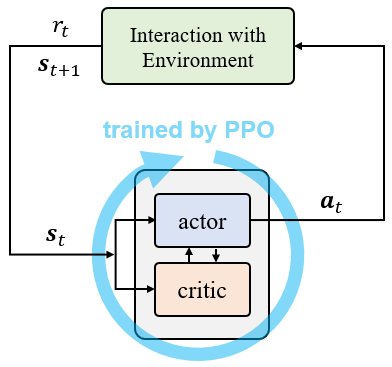}
\caption{Illustration of the overall procedure of the proposed PPO-based automatic berthing system.}
\label{fig:overall_procedure_PPO}
\end{figure}

\begin{figure}[h]
\centering
\includegraphics[width=0.25\columnwidth]{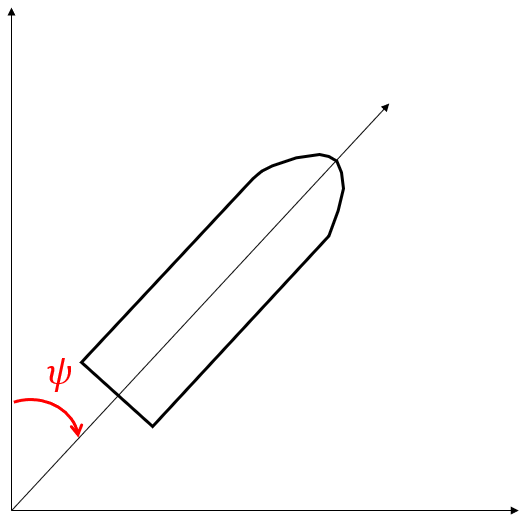}
\caption{Direction of the heading angle $\psi$ where the positive direction is indicated by the arrow.}
\label{fig:dir_heading_ang}
\end{figure}

In Algorithm \ref{alg:overall_procedure_PPO}, the overall procedure of the proposed PPO-based automatic berthing system is presented. Its corresponding illustration is shown in Fig. \ref{fig:overall_procedure_PPO}. When setting the initial ship position randomly, it should be made sure that minimum and maximum values for the random-initial ship position are defined in advance so that the random-initial ship positions can always be set within the defined range. The ship position consists of ship positions in $x$ and $y$ axes in a global coordinate system and ship’s heading angle. The configuration of the state $s_t$ is presented in Eqs. (\ref{eq:s_t})-(\ref{eq:d_t}) where $\eta$ and $\xi$ are normalized ship’s $x$, $y$ positions by length of a ship $L$. $\psi$ denotes the heading angle and its direction is presented in Fig. \ref{fig:dir_heading_ang}. $d$ is distance between a berthing goal point $g$ and the normalized ship position. $u$, $v$, and $r$ denote velocities in the surge, sway, yaw directions, respectively. The subscript $t$ denotes timestep. The action $a_t$ the agent takes consists of the rudder angle and RPS as shown in Eq. (\ref{eq:a_t}) in which $\delta$ and $n$ denote the rudder angle and RPS, respectively. The reward $r$ is calculated by a reward function shown in Algorithm \ref{alg:reward_func} where a unit of the rudder angle $\delta$ is degree and the tolerance can be set considering how closely the ship should berth at the berthing goal point. $\psi^{\prime}$ is a local heading angle that is zero when a ship's front is directly headed towards $g$. Finally, the training of the agent with the PPO is conducted every $N$ steps and a number of steps $N$ is determined by how much previous data to feed to train the agent at every training process. 

\begin{algorithm}[h]
\caption{Overall procedure of the proposed PPO-based automatic berthing system.}
\label{alg:overall_procedure_PPO}
\begin{algorithmic}[1] 
\STATE Initialize actor and critic.
\FOR{$\mathrm{episode} = 1, 2, \cdots$}
    \STATE Randomly select an initial ship position $\{x_0, y_0, \psi_0\}$
    \STATE Set an initial state $s_0$.
    \FOR{$t = 1, 2, 3, 3000$}
        \STATE Take action $a_t$ by the actor.
        \STATE Interact with the environment $\rightarrow r_t, s_{t+1}$
        \STATE Train the actor and critic at every $N$ step.
    \ENDFOR
\ENDFOR
\end{algorithmic}
\end{algorithm}

\begin{equation} 
s_t = \{ \eta_t, \xi_t, d_t, \psi_t, u_t, v_t, r_t \}
\label{eq:s_t}
\end{equation}
\begin{equation} 
\eta_t = x_t / L
\label{eq:eta_t}
\end{equation}
\begin{equation} 
\xi_t = y_t / L
\label{eq:xi_t}
\end{equation}
\begin{equation} 
d_t = \sqrt{(g_x - \eta_t)^2 + (g_y - \xi_t)^2}
\label{eq:d_t}
\end{equation}
\begin{equation} 
a_t = \{ \delta_t, n_t \}
\label{eq:a_t}
\end{equation}

\begin{algorithm}[h]
\caption{Reward function.}
\label{alg:reward_func}
\textbf{Input}: $d_t, \psi^{\prime}_t, \delta_t, u_t$\\
\textbf{Parameter}: $\mathrm{tolerance}$\\
\textbf{Output}: $r_t$
\begin{algorithmic}[1] 
\STATE $r_t = 0$
\IF{$d_t \leq \mathrm{tolerance}$}
    \STATE $r_t = r_t + 10$
    \IF{$-15^\circ \leq \psi^{\prime}_t \leq 15^\circ$}
        \STATE // the above condition encourages the ship to be headed towards $g$.
        \STATE $r_t = r_t + 2$
    \ENDIF
\ENDIF
\STATE
\STATE $r_t = r_t - \mathrm{abs}(\delta_t) / 500$  \quad// to prevent an excessive rudder control.
\STATE
\IF{$u < 0$}
    \STATE $r_t = r_t + u_t/10$  \quad// to encourage going forward.
\ENDIF

\STATE
\STATE $r_t = r_t/10$  \quad// scaling

\end{algorithmic}
\end{algorithm}

\section{Architecture of the Agent}
\begin{figure}[h]
\centering
\includegraphics[width=0.3\columnwidth]{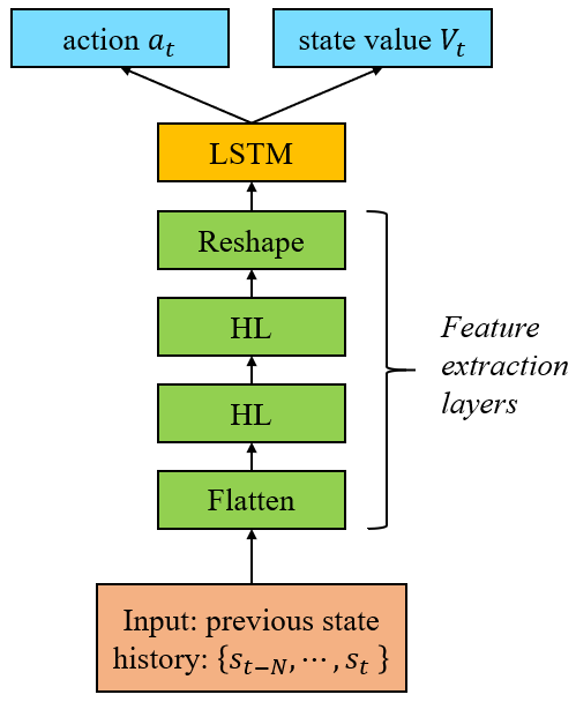}
\caption{Architecture of the agent.}
\label{fig:architecture_of_agent}
\end{figure}

The architecture of the agent is shown in Fig. \ref{fig:architecture_of_agent}. The length of the previous state history is determined by $N$ which is shown in Algorithm \ref{alg:overall_procedure_PPO}. The boxes of the Flatten, HL, Reshape, LSTM denote a flatten layer, hidden layer, reshape layer, and an LSTM layer, respectively. The feature extraction layers extract embedded features from the previous state history and the extracted features are fed into the LSTM layer. The state value $V_t$ is one of the agent’s output and it measures how good the current state is considering future rewards the agent may receive. This state value is used to train the agent with the PPO to learn the optimal policy that can take appropriate actions to maximize the reward received throughout an episode. In this study, the size of the HL and LSTM is set to 64 and 256, respectively.

\section{Simulation Conditions}
Principal particulars of a target ship are shown in Table \ref{tab:ship_property}. Control restrictions on the propeller and the rudder are shown in Eqs. (\ref{eq:n_range})-(\ref{eq:derivative_delta_range}) where units of $n$, $\delta$, and $\left| {d\delta}/{dt} \right|$ are RPS (rotation per second), deg, and deg/s, respectively. There is no restriction on a change rate of $n$ in this paper in order to first check feasibility of application of PPO on the automatic berthing system. No environmental load is considered such as wind force to put a focus on learning capability and berthing performance of the proposed PPO-based automatic berthing system when there is no environmental disturbance. The ranges of the random-initial ship positions $\eta_0$, $\xi_0$, and $\psi_0$ in Algorithm \ref{alg:overall_procedure_PPO} are presented in Eqs. (\ref{eq:eta_0 range})-(\ref{eq:psi_prime_0 range}) where the subscript 0 denotes a timestep of zero. The initial heading angle $\psi_0$ is basically set towards the berthing goal point from the initial ship position with some value added which is sampled from a uniform distribution $U(-15^{\circ}, 15^{\circ})$. The berthing goal point is defined as in Eq. (\ref{eq:gxgy}). The number of steps $N$ in Algorithm \ref{alg:overall_procedure_PPO} is set to 128. The hyperparameter settings from the original PPO paper are used with a bit of change.

\begin{table}[h]
\centering
\caption{Principal particulars of a target ship.}
\label{tab:ship_property}
\begin{tabular}{lc}
\hline
\multicolumn{2}{c}{\textbf{HULL}}\xrowht[()]{5pt}      \\
\hline
Length overall                & 188 m  \\
Length between perpendiculars & 175 m  \\
Breath                        & 25.4 m \\
Draft                         & 8.5    \\
Block coefficient             & 0.559  \\
\hline
\multicolumn{2}{c}{\textbf{RUDDER}}\xrowht[()]{5pt}    \\
\hline
Height                        & 7.7 m  \\
Area ratio                    & 1/45.8 \\
Aspect ratio                  & 1.827  \\
\hline
\multicolumn{2}{c}{\textbf{PROPELLER}}\xrowht[()]{5pt} \\
\hline
Diameter                      & 6.5 m  \\
Pitch ratio                   & 1.055  \\
Expanded area ratio           & 0.73   \\
\hline
\end{tabular}
\end{table}

\begin{equation} 
1 \leq n \leq 1
\label{eq:n_range}
\end{equation}
\begin{equation} 
-35 \leq \delta \leq 35
\label{eq:delta_range}
\end{equation}
\begin{equation} 
0 \leq \left| d\delta/dt \right| \leq 3
\label{eq:derivative_delta_range}
\end{equation}

\begin{equation} 
7 \leq \eta_0 \leq 12
\label{eq:eta_0 range}
\end{equation}
\begin{equation} 
2 \leq \xi_0\leq 9
\label{eq:xi_0 range}
\end{equation}
\begin{equation} 
\psi^{\prime}_0 - 15^\circ \leq \psi_0 \leq \psi^{\prime}_0 + 15^\circ
\label{eq:psi_prime_0 range}
\end{equation}
\begin{equation} 
g_x, g_y = [1.5, 1.5]
\label{eq:gxgy}
\end{equation}

\section{Simulation and Result Discussion}
\begin{figure}[t]
\centering
\includegraphics[width=0.5\columnwidth]{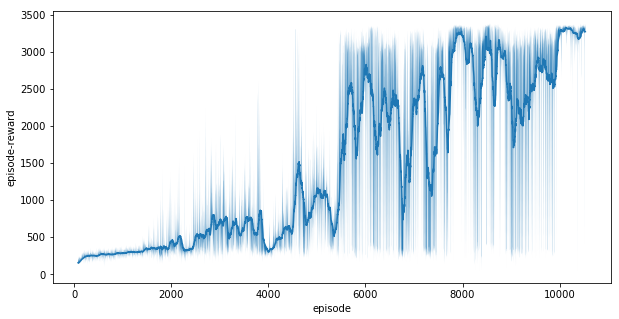}
\caption{Reward time history of training the PPO-based automatic berthing system.}
\label{fig:reward_time_hist}
\end{figure}

\begin{figure*}[ht!]
\centering
\includegraphics[width=0.9\textwidth]{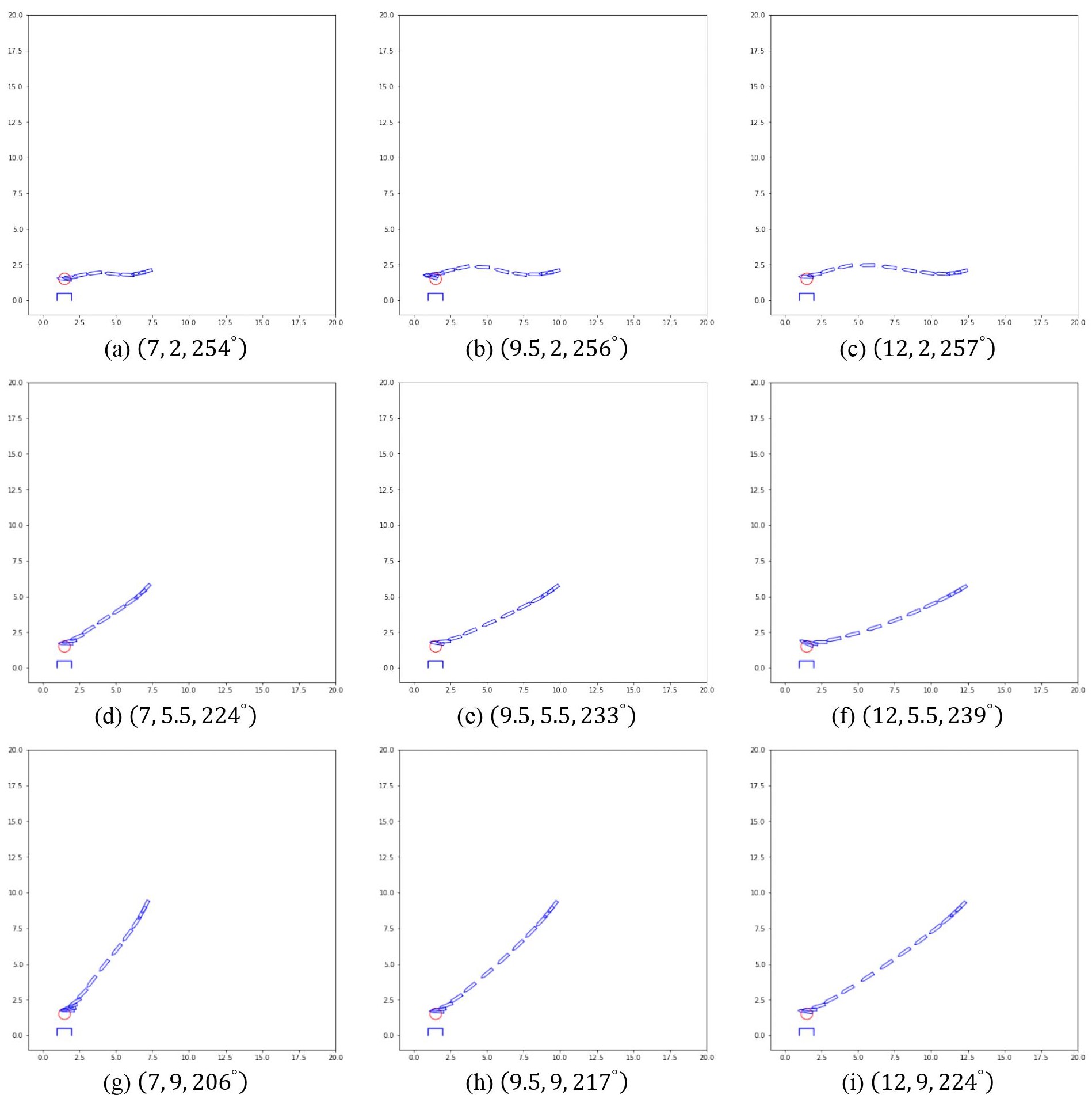}
\caption{Berthing trajectories by the proposed PPO-based automatic berthing system
(figure sub-index) denotes ($\eta_0$, $\xi_0$, $\psi_0$). The initial positions are within the random-initial ship position range during training (\textit{interpolated}).}
\label{fig:berthing_trajectory_PPO_interpolated}
\end{figure*}

\begin{figure*}[ht!]
\centering
\includegraphics[width=0.9\textwidth]{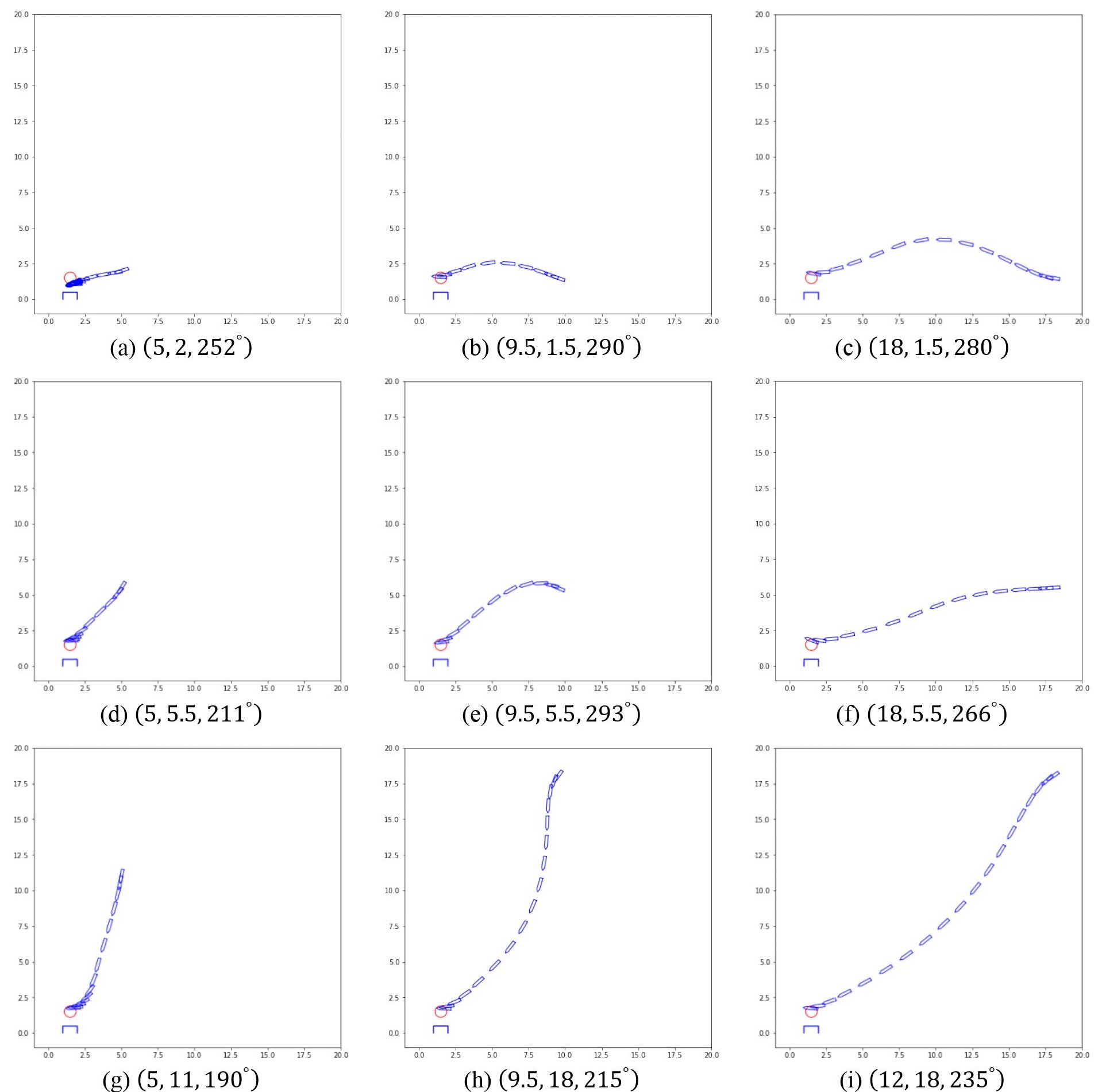}
\caption{Berthing trajectories by the proposed PPO-based automatic berthing system
(figure sub-index) denotes ($\eta_0$, $\xi_0$, $\psi_0$). The initial positions are outside of the random-initial ship position range during training (\textit{extrapolated}).}
\label{fig:berthing_trajectory_PPO_extrapolated}
\end{figure*}

Training progress of the RL algorithm can be observed through a reward time history and its convergence. Since the goal of the RL algorithm is to find the optimal policy which can maximize rewards received throughout an episode, the training of the RL algorithm is finished when the reward time history increased and finally converges. The proposed PPO-based automatic berthing system was trained for about 20 hours and its reward time history is shown in Fig. \ref{fig:reward_time_hist} where the reward time history is filtered by a moving average filter with a window size of 0.99. In Fig. \ref{fig:reward_time_hist}, it can be observed that the reward time history increases in an early stage and converges later, meaning that the PPO reached the optimal policy when converged. Next, berthing trajectories using the trained PPO-based automatic berthing system are presented in Fig. \ref{fig:berthing_trajectory_PPO_interpolated} and Fig. \ref{fig:berthing_trajectory_PPO_extrapolated} where the red circle is the berthing goal point given the tolerance, the blue half-square denotes a harbor, and the ship in the trajectory is drawn every 50s. The control time histories of the propeller and the rudder angle for Fig. \ref{fig:berthing_trajectory_PPO_extrapolated} can be found in Figs. \ref{fig:time_series_berthing_trajectory_PPO_extrapolated(a)}-\ref{fig:time_series_berthing_trajectory_PPO_extrapolated(i)}. Fig. \ref{fig:berthing_trajectory_PPO_interpolated} shows that the berthing control is well managed for the initial ship positions within the random initial ship position range during training, which can be viewed as control within the interpolated range. More importantly, Fig. \ref{fig:berthing_trajectory_PPO_extrapolated} shows that the berthing control is successful and robust even for the initial ship positions that are outside of the training initial ship position range, which can be viewed as control starting from the \textit{extrapolated} range. Therefore, it can be argued that PPO can learn the optimal control policy that is generic enough to perform well even in extrapolated situations, and the further study is encouraged to be conducted with the additional restriction on $n$ and environmental load.

Code for the simulation is available here\footnote{\texttt{https://github.com/danelee2601/RL-based-automatic-berthing}}. In the GitHub repository, a link to the Google Colab is available where you can run the berthing simulation with the pre-trained model by the PPO. There, you can try various initial ship positions and would be able to see the robustness of the PPO-based automatic berthing system.

\section{Conclusion}
The PPO-based automatic berthing system is proposed in this study. The main limitation of the previous ANN-based automatic berthing system comes from having to have a large number of berthing data as training data required to train the ANN. However, the berthing data is difficult to obtain. There are three main advantages of the proposed PPO-based automatic berthing system in comparison to the previous ANN-based automatic berthing system. First, the proposed system does not require any pre-obtained training data because it obtains data through interaction with a given environment. Second, the proposed system is not limited by having to have a large number of training data because it can obtain the training data indefinitely as long as simulation continues. Third, with a careful definition of the reward function, a desired maneuvering behavior can be achieved without any pre-knowledge in the ship maneuvering. In the simulation results, it was shown that the proposed system could learn how to robustly maneuver the ship to the berthing goal point from a variety of the initial positions.

\bibliographystyle{unsrt}
\bibliography{main}

\newpage
\appendix
\begin{figure*}[ht!]
\centering
\includegraphics[width=0.7\textwidth]{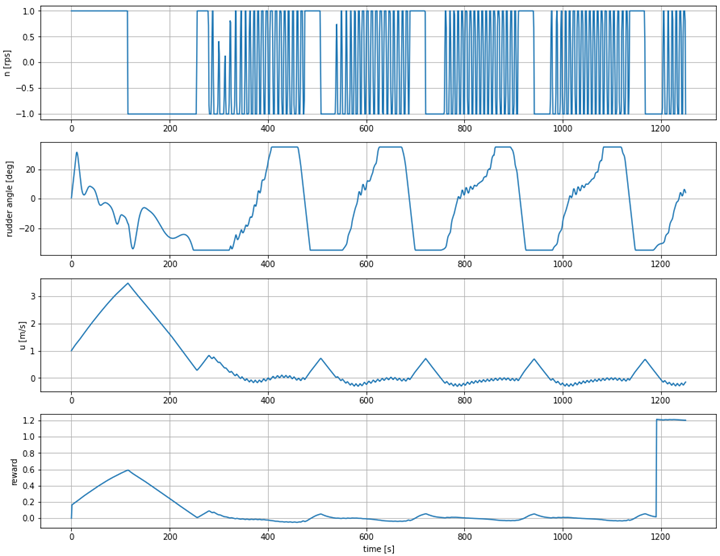}
\caption{Time series of $n$, $\delta$, $u$, and reward of Fig. \ref{fig:berthing_trajectory_PPO_extrapolated}(a)}
\label{fig:time_series_berthing_trajectory_PPO_extrapolated(a)}
\end{figure*}

\begin{figure*}[ht!]
\centering
\includegraphics[width=0.7\textwidth]{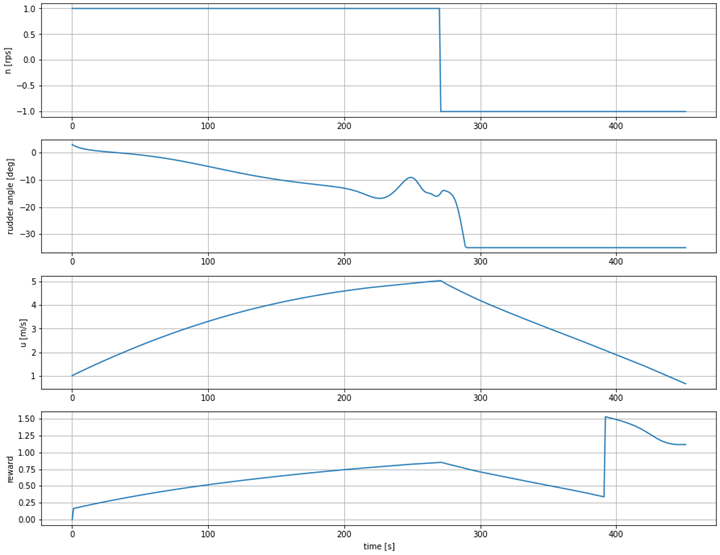}
\caption{Time series of $n$, $\delta$, $u$, and reward of Fig. \ref{fig:berthing_trajectory_PPO_extrapolated}(b)}
\end{figure*}

\begin{figure*}[ht!]
\centering
\includegraphics[width=0.7\textwidth]{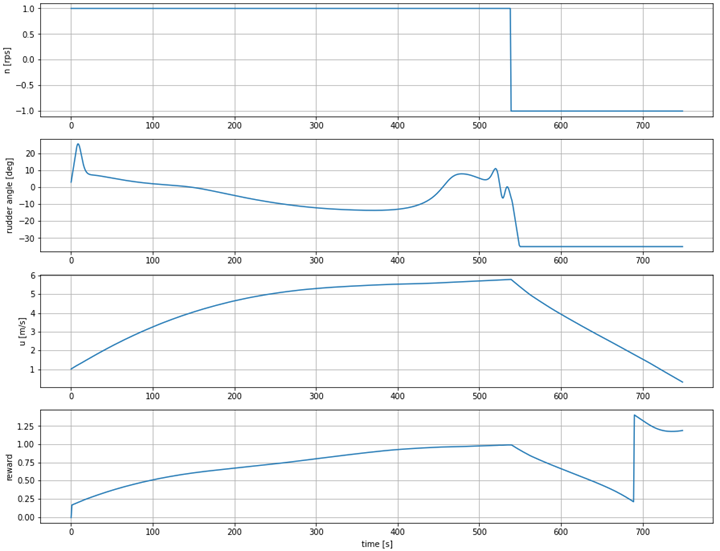}
\caption{Time series of $n$, $\delta$, $u$, and reward of Fig. \ref{fig:berthing_trajectory_PPO_extrapolated}(c)}
\end{figure*}

\begin{figure*}[ht!]
\centering
\includegraphics[width=0.7\textwidth]{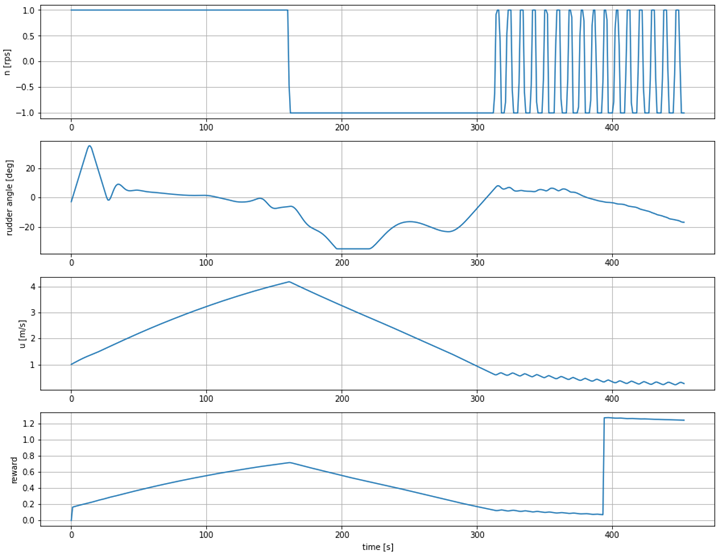}
\caption{Time series of $n$, $\delta$, $u$, and reward of Fig. \ref{fig:berthing_trajectory_PPO_extrapolated}(d)}
\end{figure*}

\begin{figure*}[ht!]
\centering
\includegraphics[width=0.7\textwidth]{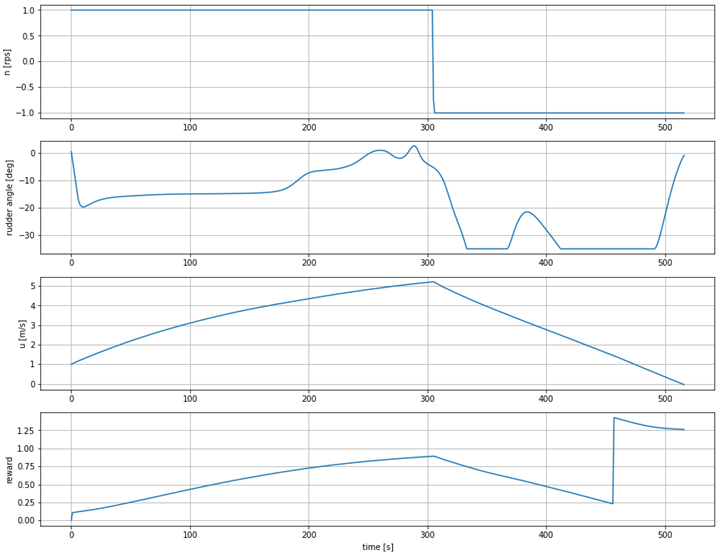}
\caption{Time series of $n$, $\delta$, $u$, and reward of Fig. \ref{fig:berthing_trajectory_PPO_extrapolated}(e)}
\end{figure*}

\begin{figure*}[ht!]
\centering
\includegraphics[width=0.7\textwidth]{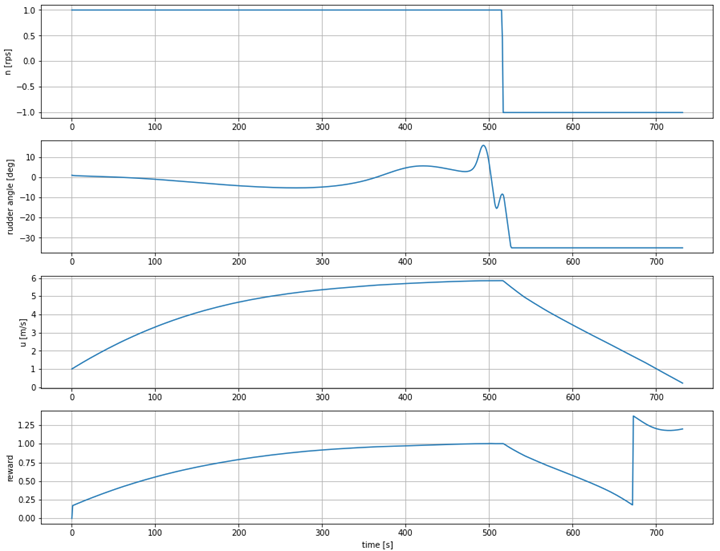}
\caption{Time series of $n$, $\delta$, $u$, and reward of Fig. \ref{fig:berthing_trajectory_PPO_extrapolated}(f)}
\end{figure*}

\begin{figure*}[ht!]
\centering
\includegraphics[width=0.7\textwidth]{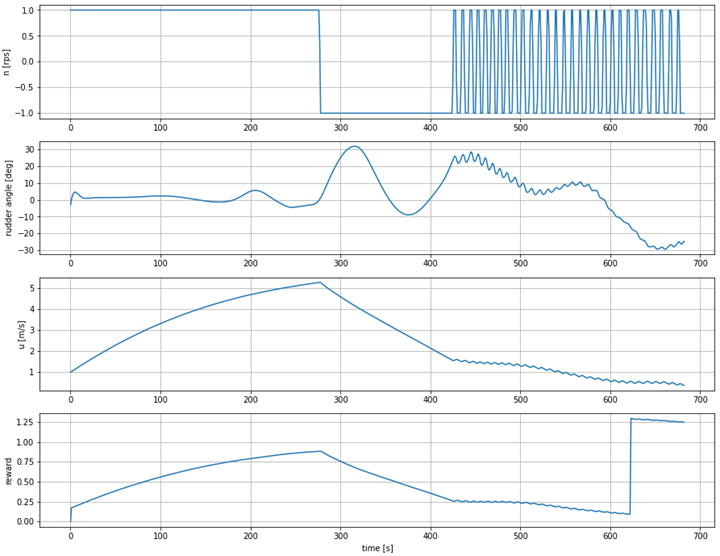}
\caption{Time series of $n$, $\delta$, $u$, and reward of Fig. \ref{fig:berthing_trajectory_PPO_extrapolated}(g)}
\end{figure*}

\begin{figure*}[ht!]
\centering
\includegraphics[width=0.7\textwidth]{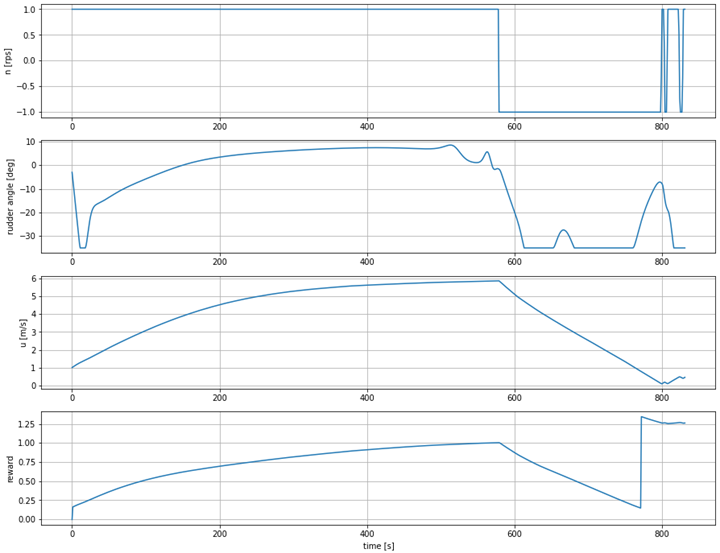}
\caption{Time series of $n$, $\delta$, $u$, and reward of Fig. \ref{fig:berthing_trajectory_PPO_extrapolated}(h)}
\end{figure*}

\begin{figure*}[ht!]
\centering
\includegraphics[width=0.7\textwidth]{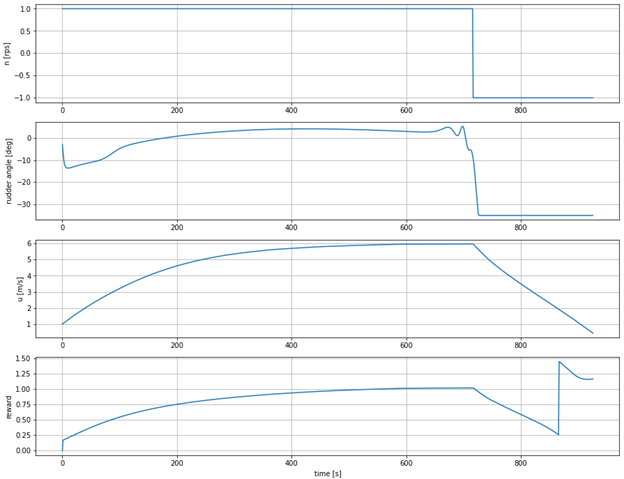}
\caption{Time series of $n$, $\delta$, $u$, and reward of Fig. \ref{fig:berthing_trajectory_PPO_extrapolated}(i)}
\label{fig:time_series_berthing_trajectory_PPO_extrapolated(i)}
\end{figure*}

\end{document}